\documentclass[letterpaper,conference]{IEEEtran}
\IEEEoverridecommandlockouts
\usepackage{graphics} 
\usepackage[linesnumbered,ruled]{algorithm2e}
\graphicspath{{img/}}
\usepackage{epsfig} 
\usepackage{amsmath} 
\usepackage{amssymb}  
\usepackage{amsthm}
\usepackage{bigints}
\usepackage{pifont}
\def\-{\raisebox{.75pt}{-}}
\usepackage{mathtools}
\usepackage{cite}
\usepackage[makeroom]{cancel}
\usepackage{xcolor} 

\usepackage[center]{subfigure}
\usepackage{multirow}
\usepackage{hyperref}
\usepackage{booktabs}
\usepackage{subfigure}
\usepackage{epstopdf}
\usepackage{textcomp}
\usepackage{bm}
\usepackage{caption}
\usepackage{algpseudocode,algorithm} 
\usepackage{color}
\usepackage{mathptmx}
\usepackage[11pt]{moresize}
\usepackage{hhline}
\usepackage{tikz}
\usepackage{textcomp}
\def\BibTeX{{\rm B\kern-.05em{\sc i\kern-.025em b}\kern-.08em
    T\kern-.1667em\lower.7ex\hbox{E}\kern-.125emX}}

\newcommand\copyrighttext{%
\centering \footnotesize \copyright 2019 IEEE, to appear in the 30th IEEE Intelligent Vehicles Symposium. }
\newcommand\copyrightnotice{%
\begin{tikzpicture}[remember picture,overlay]
\node[anchor=south,yshift=10pt] at (current page.south) {\fbox{\parbox{\dimexpr\textwidth-\fboxsep-\fboxrule\relax}{\copyrighttext}}};
\end{tikzpicture}%
}

\begin{document}
\title{\LARGE Leveraging Heteroscedastic Aleatoric Uncertainties for Robust Real-Time LiDAR 3D Object Detection
\thanks{$^1$ Robert Bosch GmbH, Corporate Research, Driver Assistance Systems and Automated Driving, 71272 Renningen, Germany.}
\thanks{$^2$ Institute of Measurement, Control and Microtechnology, Ulm University, 89081 Ulm, Germany.}
\thanks{The video to this paper can be found at \url{https://youtu.be/2DzH9COLpkU}.}
}
\author{Di Feng$^{1,2}$, Lars Rosenbaum$^1$, Fabian Timm$^1$, Klaus Dietmayer$^2$}

\maketitle

\begin{abstract}
We present a robust real-time LiDAR 3D object detector that leverages heteroscedastic aleatoric uncertainties to significantly improve its detection performance. A multi-loss function is designed to incorporate uncertainty estimations predicted by auxiliary output layers. Using our proposed method, the network ignores to train from noisy samples, and focuses more on informative ones. We validate our method on the KITTI object detection benchmark. Our method surpasses the baseline method which does not explicitly estimate uncertainties by up to nearly $9\%$ in terms of Average Precision (AP). It also produces state-of-the-art results compared to other methods, while running with an inference time of only $72ms$. In addition, we conduct extensive experiments to understand how aleatoric uncertainties behave. Extracting aleatoric uncertainties brings almost no additional computation cost during the deployment, making our method highly desirable for autonomous driving applications. 
\end{abstract}

\copyrightnotice

\begin{figure*}[tbp]
	\centering
	\begin{minipage}{1\textwidth}
		\centering
		\includegraphics[width=0.96\linewidth]{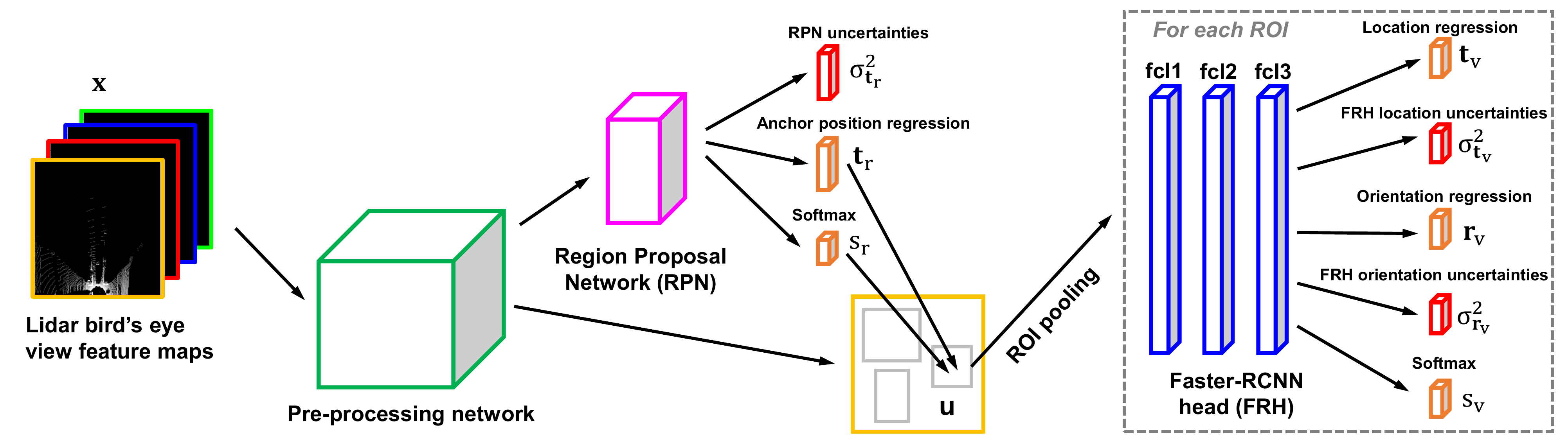}
	\end{minipage}
	\caption{Our proposed LiDAR 3D object detector which predicts the heteroscedastic aleatoric uncertainties. The network takes the LiDAR bird's eye view maps as input and predicts softmax probability, object 3D bounding boxes and orientations. The network consists of three parts: a pre-processing network based on the VGG16 to extract high-level LiDAR features; a Region Proposal Network (RPN) that produces Region of Interests (ROI); a Faster-RCNN head (FRH) which fine-tunes each ROI. The network regresses the aleatoric uncertainties $\sigma^2_{\textbf{t}_r}$, $\sigma^2_{\textbf{t}_v}$ and $\sigma^2_{\textbf{r}_v}$ in RPN and FRH via auxiliary output layers.}\label{fig:network}
    \vspace{-0.5em}
\end{figure*}

\setlength{\parskip}{2mm plus3mm minus3mm}
\setlength{\belowdisplayskip}{8pt} \setlength{\belowdisplayshortskip}{8pt}
\setlength{\abovedisplayskip}{8pt} \setlength{\abovedisplayshortskip}{8pt}

\section{Introduction}\label{sec:introduction}
A robust and accurate object detection system using on-board sensors (e.g. camera, LiDAR, radar) is crucial for the road scene understanding of autonomous driving. Among different sensors, LiDAR can provide us with accurate depth information, and is robust under different illumination conditions such as daytime and nighttime. These properties make LiDAR indispensable for safe autonomous driving. The recent Uber's autonomous driving fatal tragedy could have been avoided, if the LiDAR perception system had robustly detected the pedestrian, or had timely informed the human driver to trigger the emergency braking because it was uncertain with the driving situation~\cite{ntsb_2018}.

Recently, deep learning approaches have brought significant improvement to the object detection problem~\cite{janai2017computer}. Many methods have been proposed that use LiDAR point clouds ~\cite{li20163d,li2016vehicle,zhou2017voxelnet, engelcke2017vote3deep}, or fuse them with camera images~\cite{chen2016multi,ku2017joint,xu2017pointfusion,qi2017frustum,du2018general,pfeuffer2018optimal}. However, they only give us deterministic bounding box regression and use softmax scores to represent recognition probability, which does not necessarily represent uncertainties in the network~\cite{Gal2016Uncertainty}. In other words, they do not provide detection confidence regarding the classification and localization. For a robust perception system, we need to explicitly model the network's uncertainties. 

Towards this goal, in this work we build a probabilistic 2-stage object detector from LiDAR point clouds by introducing heteroscedastic aleatoric uncertainties - the uncertainties that represent sensor observation noises and vary with the data input. The method works by adding auxiliary outputs to model the aleatoric uncertainties, and training the network with a robust multi-loss function. In this way, the network learns to focus more on informative training samples and ignore the noisy ones. Our contributions can be summarized as follows: 
\begin{itemize}
\item We model heteroscedastic aleatoric uncertainties in the Region Proposal Network (RPN) and the Faster-RCNN header (FRH) of a LiDAR 3D object detector.
\item We show that by leveraging aleatoric uncertainties, the network produces state-of-the-art results and significantly increases the average precision up to $9\%$ compared to the baseline method without any uncertainty estimations.
\item We systematically study how the aleatoric uncertainties behave. We show that the uncertainties are related with each other, and are influenced by multiple factors such as the detection distance, occlusion, softmax score, and orientation.
\end{itemize}

In the sequel, we first summarize related works in Sec.~\ref{sec:works}, and then describe our proposed method in detail in Sec.~\ref{sec:method}. Sec.~\ref{sec:result} shows the experimental results regarding (1) the improvement of object detection performance by leveraging aleatoric uncertainties, and (2) an analysis on how the uncertainties behave. Sec.~\ref{sec:conclusion} summarizes the work and discusses future research. The video of this work is provided as a supplementary material.

\section{Related Works}
\label{sec:works}
In the following, we summarize methods for LiDAR-based object detection in autonomous driving and uncertainty quantification in Bayesian Neural Networks.

\subsection{LiDAR-based Object Detection} 
Many works process the LiDAR data in 3D space~\cite{zhou2017voxelnet,engelcke2017vote3deep,li20163d,xu2017pointfusion,qi2017frustum}. For example, Zhou \textit{et al.}~\cite{zhou2017voxelnet} propose a voxel feature encoding layer to handle discretized 3D point clouds. Li~\cite{li20163d} employs a 3D fully convolutional neural network to predict an objectness map and a 3D bounding box map. Xu \textit{et al.}~\cite{xu2017pointfusion} propose to directly process 3D LiDAR points without discretization via the PointNet~\cite{qi2017pointnet}. Besides, other works project 3D point clouds onto a 2D plane and use the 2D convolutional network to process  LiDAR feature maps, such as front-view cylindrical images~\cite{li2016vehicle,wu2017squeezeseg,chen2016multi}, camera-coordinate images~\cite{kim2016robust,schlosser2016fusing}, and bird's eye view (BEV) maps~\cite{feng2018towards,caltagirone2017fast,Yang_2018_CVPR}. In addition to the LiDAR sensors, an autonomous driving car is usually equipped with other sensors such as cameras or radar sensors. Therefore, it is natural to fuse them for more robust and accurate object detection. For instance, Chen \textit{et al.}~\cite{chen2016multi} use LiDAR BEV maps to generate region proposals, and fuse the regional features from LiDAR BEV and front-view maps, as well as camera images for 3D car detection. Qi \textit{et al.}~\cite{qi2017frustum} propose to generate 2D bounding boxes by an image object detector, and use the regional point clouds within these bounding boxes for 3D object detection. In this work, we build a LiDAR-only object detector using BEV representations. The network estimates 3D bounding boxes and object orientations.

\subsection{Uncertainty Quantification in Bayesian Neural Networks} 
A common way to estimate probabilities in deep neural networks is by Bayesian Neural Networks (BNNs), which assume a prior distribution of network weights and conduct inference in the posterior distribution over them~\cite{mackay1992practical}. There are two types of uncertainties we can model in BNNs~\cite{kendall2017uncertainties}, namely, the \textit{epistemic} and the \textit{aleatoric} uncertainty. The \textit{epistemic} uncertainty shows the model's uncertainty when describing the training dataset. It can be quantified through variational inference~\cite{hinton1993keeping} or the sampling techniques~\cite{graves2011practical,Gal2016Uncertainty}, and has been applied to active learning~\cite{gal2017deep,Beluch_2018_CVPR}, image semantic segmentation~\cite{kampffmeyer2016semantic,kendall2015bayesian}, camera location estimation~\cite{kendall2015modelling} or open-dataset object detection problems~\cite{miller2017dropout}. The \textit{aleatoric} uncertainty, on the other hand, models the observation noises of input data. It has been modeled for geometry and semantic predictions in camera images~\cite{kendall2017uncertainties, kendall2017multi}. Recently, Ilg~\textit{et al.}~\cite{ilg2018uncertainty} extract epistemic and aleatoric uncertainties in an optical flow problem, and we study how these uncertainties distinguish from each other in an object detector~\cite{feng2018towards}. Unlike our previous method, in this work we focus on modelling the aleatoric uncertainties both in RPN and FRH, and systematically study how it improves the network's robustness against the noisy LiDAR data.
\section{Methodology}
\label{sec:method}
In this section, we present our method which leverages heteroscedastic aleatoric uncertainties for the robust LiDAR 3D object detection. We start with illustrating our network architecture, which is followed by a description of how to model the uncertainties. We end the section by introducing a robust multi-loss function that enables the network to learn the aleatoric uncertainties. The network architecture is shown in Fig.~\ref{fig:network}.
\subsection{Network Architecture}
\label{subsec:architecture}
\subsubsection{LiDAR Point Clouds Transformation}
In this work, we follow~\cite{chen2016multi} to project the LiDAR 3D point clouds onto 2D grids, and use the bird's eye view (BEV) feature maps as network inputs. These LiDAR feature maps comprise several height maps and a density map. To obtain height maps, we first divide point clouds along the vertical axis of the LiDAR coordinate frame ($z$ axis) into several slices. We then discretize each slice into 2D grid cells, and encode the height information as the maximum height values of points within a cell. As for the density map, we assign the value $min(1,\frac{\log(N+1)}{\log(16)})$ in each cell, where $N$ refers to the number of points in a cell.
\subsubsection{Two-stage Object Detector}
We follow the two-stage object detection network proposed in~\cite{ren2015faster}. The LiDAR BEV feature maps are fed into a pre-processing network based on VGG16 to extract high-level LiDAR features. 
After the pre-processing layers, a region proposal network (RPN) produces 3D region of interests (ROIs) based on pre-defined 3D  anchors at each pixel on the feature map. The RPN consists of two task-specific fully connected layers, each with $256$ hidden units. A ROI is parametrized by $[x_{r},y_{r},z_{r},w_{r},l_{r},h_{r},s_{r}]$, with $x_{r}$ and $y_{r}$ indicating the ROI position in the bird's eye view plane, $z_{r}$ its height, $w_{r}$, $l_{r}$, $h_{r}$ its dimensions, and $s_{r}$ the softmax objectness score (``Object" or ``Background"). The anchor's dimensions are determined by clustering the training samples into two clusters via the k-means algorithm. Its height is defined as the LiDAR's height above the ground plane. Similar to~\cite{ren2015faster}, the RPN regresses the region proposals by the normalized offsets denoted as $\mathbf{t}_r = [\Delta x_{r}, \Delta y_{r}, \Delta z_{r}, \Delta w_{r}, \Delta l_{r}, \Delta h_{r}]$ and predicts the objectness score $s_{r}$ by a softmax layer. 

The Faster-RCNN head (FRH) with three fully connected hidden layers ($2048$ units for each layer) is designed to fine-tune ROIs generated by RPN. It produces multi-task outputs, including the softmax probability of the object classes and the background $s_{v}$, the 3D bounding box location, and the object orientation. We encode the location with four corner method introduced in~\cite{ku2017joint}: $ \mathbf{t}_v = [\Delta x_{v}^{(1)} ... \Delta x_{v}^{(4)};\ \  \Delta y_{v}^{(1)} ... \Delta y_{v}^{(4)};\ \  \Delta h_{v}^{(1)}, \Delta h_{v}^{(2)}]$, with $\Delta x_{v}$ and $\Delta y_{v}$ being the relative position of a bounding box corner in $x$, $y$ axes of the LiDAR coordinate frame, and $\Delta h_{v}^{(1)}, \Delta h_{v}^{(2)}$ being the height offsets from the ground plane. We also encode the orientation as $\mathbf{r}_v = [\cos(\theta), \sin(\theta)]$, with $\theta$ being the object orientation in BEV. As explained in \cite{ku2017joint}, explicitly modeling the orientation can remedy angle wrapping problem, resulting in a better object detection performance.
\subsection{Modeling Heteroscedastic Aleatoric Uncertainties} \label{subsec:aleatoric}
As introduced in Sec.~\ref{sec:introduction}, the heteroscedastic aleatoric uncertainties indicate data-dependent observation noises in LiDAR sensor. For example, a distant or occluded object should yield high aleatoric uncertainties, since there are a few LiDAR points representing them. In our proposed robust LiDAR 3D object detector, we extract aleatoric uncertainties in both RPN and FRH.

Let us denote an input LiDAR BEV feature image as $\mathbf{x}$, and a region of interest produced by RPN as $\mathbf{u}$. The object detector predicts the classification labels $y$ with one-hot encoding from RPN and FRH, denoted by $y_r$ and $y_c$ respectively, i.e. $y \in \{y_r, y_c\}$. The detector also predicts the anchor positions $\mathbf{t}_r$, object locations $\mathbf{t}_v$ and orientations $\mathbf{r}_v$, denoted as $\mathbf{l}$, i.e. $\mathbf{l} \in \{\mathbf{t}_{r}, \mathbf{t}_{v}, \mathbf{r}_{v}\}$. Additionally, we refer $\mathbf{M}$ to be the network weights and $\mathbf{f}^{\mathbf{M}}(\cdot)$ the noise-free network outputs both for classification and regression problems. Specifically, for classification $\mathbf{f}^{\mathbf{M}}(\cdot)$ indicates the network output logits \textit{before} the softmax function.

We use multivariate Gaussian distributions with diagonal covariance matrices to model the observation likelihood for the regression problems $p(\mathbf{l}|\mathbf{f}^{\mathbf{M}}(\cdot))$, including the anchor position $p(\mathbf{t}_{r}|\mathbf{f}^{\mathbf{M}}(\mathbf{x}))$, the 3D bounding box location $p(\mathbf{t}_{v}|\mathbf{f}^{\mathbf{M}}(\mathbf{u}))$ and the object orientation $p(\mathbf{r}_{v}|\mathbf{f}^{\mathbf{M}}(\mathbf{u}))$:
\begin{equation}\label{equ:alea_reg}
\begin{split}
&p(\mathbf{l}|\mathbf{f}^{\mathbf{M}}(\cdot))=\mathcal{N}(\mathbf{f}^{\mathbf{M}}(\cdot)),\Sigma_{\mathbf{l}}), \ \ \ \ \ \Sigma_{\mathbf{l}}=diag(\sigma^2_{\mathbf{l}}), \\
\end{split}
\end{equation}
with $\sigma^2_{\mathbf{l}} \in \{\sigma^2_{\mathbf{t}_{r}}, \sigma^2_{\mathbf{t}_{v}}, \sigma^2_{\mathbf{r}_{v}}\}$ referring to the observation noise vectors. Each element in these vectors represents an aleatoric uncertainty scalar and corresponds to an element in $\mathbf{t}_{r}$, $\mathbf{t}_{v}$ and $\mathbf{r}_{v}$. For instance, for the object orientation prediction which is encoded by $\mathbf{r}_v = [\cos{\theta}, \sin{\theta}]$, we have  $\mathbf{\sigma^2_{\mathbf{r}_{v}}} = [\sigma^2_{\cos{\theta}}, \sigma^2_{\sin{\theta}}]$. In order to estimate those observation noises, we can add auxiliary output layers to regress their values. In practice, we regress $\log \sigma^2_{\mathbf{l}}$ to increase numerical stability and consider the positive constraints. In this way, the regression outputs of RPN that model the aleatoric uncertainties can be formulated as $\mathbf{\hat{t}}_r = [\mathbf{t}_r,\ \  \log \sigma^2_{\mathbf{t}_{r}}]$, and for FRH they are $\mathbf{\hat{t}}_v = [\mathbf{t}_v,\ \  \log \sigma^2_{\mathbf{t}_{v}}]$ and $\mathbf{\hat{r}}_v = [\mathbf{r}_v,\ \  \log \sigma^2_{\mathbf{r}_{v}}]$, as illustrated by Fig.~\ref{fig:network}.

For the classification tasks, we assume that the network noise-free outputs are corrupted by the softmax function. Thus, the observation likelihood $p(y_r|\mathbf{f}^{\mathbf{M}}(\mathbf{x}))$ and $p(y_v|\mathbf{f}^{\mathbf{M}}(\mathbf{u}))$ can be written as:
\begin{equation}\label{equ:alea_cls}
\begin{split}
&p(y|\mathbf{f}^{\mathbf{M}}(\cdot)) = s = softmax(\mathbf{f}^{\mathbf{M}}(\cdot)), \\
\end{split}
\end{equation}
with $y \in \{y_r, y_v\}$ and $s \in \{s_r, s_v\}$. Here, we do not explicitly model the aleatoric classification uncertainties, as they are self-contained from the softmax scores which follow the categorical distribution.

\subsection{Robust Multi-Loss Function} \label{subsec:loss}
In order to learn the aleatoric uncertainties, we incorporate $\sigma^2_{\mathbf{t}_{r}}$, $\sigma^2_{\mathbf{t}_{v}}$ and $\sigma^2_{\mathbf{r}_{v}}$ in a multi-loss function for training the object detector:
\begin{equation}\label{equ:objective}
\begin{split}
& \mathcal{L}(\mathbf{M}, \sigma^2_{\mathbf{t}_{r}}, \sigma^2_{\mathbf{t}_{v}},\sigma^2_{\mathbf{r}_{v}}) \\
& = \underbrace{\frac{1}{2\sigma^2_{\mathbf{t}_{r}}}\mathcal{L}(\mathbf{M},\ \mathbf{t}_r) + \log \sigma^2_{\mathbf{t}_{r}}}_\text{RPN regression loss} + \underbrace{\mathcal{L}(\mathbf{M},\ y_r)}_\text{RPN classification loss} \\
& + \underbrace{\frac{1}{2\sigma^2_{\mathbf{t}_{v}}}\mathcal{L}(\mathbf{M},\ \mathbf{t}_v) + \log \sigma^2_{\mathbf{t}_{v}}}_\text{FRH location loss} + \underbrace{\mathcal{L}(\mathbf{M},\ y_v)}_\text{FRH classification loss} \\
& + \underbrace{\frac{1}{2\sigma^2_{\mathbf{r}_{v}}}\mathcal{L}(\mathbf{M},\ \mathbf{r}_v) + \log \sigma^2_{\mathbf{r}_{v}}}_\text{FRH orientation loss},
\end{split}
\end{equation}
where $\mathcal{L}(\mathbf{M},\ \mathbf{t}_r)$, $\mathcal{L}(\mathbf{M},\ \mathbf{t}_v)$ and $\mathcal{L}(\mathbf{M},\ \mathbf{r}_v)$ are the regression loss, and $\mathcal{L}(\mathbf{M}, \mathbf{y}_{r})$ and $\mathcal{L}(\mathbf{M}, \mathbf{y}_{v})$ the classification loss. Learning the aleatoric uncertainties via this multi-loss function has two effects. First, an uncertainty score $\sigma^2_{\mathbf{l}}$ can serve as a relative weight to a sub-loss. Thus, optimizing relative weights enables the object detector to balance the contribution of each sub-loss, allowing the network to be trained more easily, as discussed in~\cite{kendall2017multi}. Second, aleatoric uncertainties can increase the network robustness against noisy input data. For an input sample with high aleatoric uncertainties, i.e. the sample is noisy, the model decreases the residual regression loss $\mathcal{L}(\mathbf{M},\ \mathbf{t}_r), \mathcal{L}(\mathbf{M},\ \mathbf{t}_v), \mathcal{L}(\mathbf{M},\ \mathbf{r}_v)$ because $\frac{1}{2\sigma^2_{\mathbf{l}}}$ becomes small. Conversely, the network is encouraged to learn from the \textit{informative} samples with low aleatoric uncertainty by increasing the residual regression loss with larger $\frac{1}{2\sigma^2_{\mathbf{l}}}$ term.   

\subsection{Implementation Details}
We jointly train the RPN network and the Faster-RCNN head in an end-to-end fasion. The background anchors are ignored for the RPN regression loss. An anchor is assigned to be an object class when its Intersection over Union (IoU) with the ground truth in the BEV is larger than $0.5$, and the background class if it is below $0.3$. Anchors that are neither object nor background do not contribute to the learning. Besides, we apply Non-Maximum Suppression (NMS) with the threshold $0.8$ on the region proposals to reduce redundancy, and keep $1024$ proposals with the highest classification scores during the training process. We reduce the region proposals to $300$ during the test time in order to speed up the inference. When training the Faster R-CNN head, a ROI is labeled to be positive when its 2D IoU overlap with ground truth in BEV is larger than $0.65$, and negative when it is less than $0.55$, suggested by~\cite{ku2017joint}. We use the cross entropy loss for the classification problems, and the smooth $L_1$ loss for the regression problems, as it is more preferable than the $L_2$ loss in object detection tasks~\cite{ren2015faster}. Correspondingly, we assume the observation likelihood follows the Laplacian distribution instead of Gaussian (Eq.~\ref{equ:alea_reg}), similar to~\cite{kendall2017uncertainties}. We train the network with Adam optimizer. The learning rate is initialized as $10^{-4}$ and decayed exponentially for every $30000$ steps with a decay factor $0.8$. We also use Dropout and $l_2$ regularization to prevent over-fitting. We set the dropout rate to be $0.5$, and the weight decay to be $5\times 10^{-4}$. We first train the network without aleatoric uncertainties for $30000$ steps and then use the robust multi-loss function (Eq.~\ref{equ:objective}) for another $90000$ steps. We find that the network converges faster following this training strategy.
\section{Experimental Results}
\label{sec:result}
\subsection{Experimental Setup}
\subsubsection{Dataset}
We evaluate the performance of our proposed method on the ``Car'' category from the KITTI object detection benchmark~\cite{geiger2012we}. KITTI provides $7,481$ frames of data for training and another $7,518$ frames for testing (\textit{test} set). The dataset is classified into \textit{Easy}, \textit{Moderate}, and \textit{Hard} settings. When evaluating our method on the \textit{test} set, we train the network with all training data. As the ground truth for the \textit{test} set is not accessible, we conduct the ablation study and extensive experiments for analyzing the aleatoric uncertainties on the training dataset. To this end, we follow~\cite{chen2016multi} to split the training data into a \textit{train} set and a \textit{val} set (approximately $50/50$ split). The network is trained on the \textit{train} set and tested on the \textit{val} set.

\subsubsection{Input Representation}
We use the LiDAR point cloud within the range $length \times width \times height = [0,70] \times [$-$40,40] \times [0,2.5]$ meters in the LiDAR coordinate frame. The point clouds are discretized into $5$ height slices along the $z$ axis with $0.5$ meters resolution and the length and width are discretized with $0.1$ meters resolution, similar to~\cite{ku2017joint}. After incorporating a density feature map, the input LiDAR point clouds are represented by the feature maps with size $700 \times 800 \times 6$. 

\begin{table}[tbp]
	\centering
    \resizebox{0.96\linewidth}{!}{\begin{tabular}{|c|c c c|c c c|}
    \hline
 \multirow{2}{*}{Method} & \multicolumn{3}{|c|}{$AP_{3D} (\%)$} & \multicolumn{3}{|c|}{$AP_{BEV} (\%)$}\\ \cline{2-7}
   & Easy & Moderate & Hard & Easy & Moderate & Hard \\ \hline  
   3D FCN \cite{li20163d} & - & - & - & 69.94 & 62.54 & 55.94 \\  
    MV3D (BV+FV) \cite{chen2016multi} & 66.77 & 52.73 & 51.31 & 85.82 & 77.00 & 68.94 \\ 
    PIXOR \cite{Yang_2018_CVPR} & - & - & - & 81.70 & 77.05 & 72.95 \\
    VoxelNet \cite{zhou2017voxelnet} & 77.47 & 65.11 & 57.73 & 89.35 & 79.26 & 77.39 \\ \hline
    Baseline & 70.46 & 55.48  & 54.43 & 84.94 & 76.70 & 69.03 \\ 
	\textit{Ours} & 73.92 & 62.63 & 56.77 & 85.91 & 76.06 & 68.75  \\ \hline
    \end{tabular}}
    \caption{Comparison with state-of-the-art methods on KITTI \textit{test} set. Compared to the baseline method, \textit{Ours} simply models the aleatoric uncertainties in both RPN and FRH, without further tuning any hyper-parameters.} \label{tab:test_results}
    \vspace{-0.7em}
 \end{table}
\begin{table}[tbp]
	\centering
    \resizebox{0.96\linewidth}{!}{\begin{tabular}{|c|c c c|c c c|}
    \hline
 \multirow{2}{*}{Method} & \multicolumn{3}{|c|}{$AP_{3D} (\%)$} & \multicolumn{3}{|c|}{$AP_{BEV} (\%)$}\\ \cline{2-7}
   & Easy & Moderate & Hard & Easy & Moderate & Hard \\ \hline
    VeloFCN \cite{li2016vehicle} & 15.20 & 13.66& 15.98& 40.14& 32.08& 30.47 \\ 
    MV3D (BV+FV) \cite{chen2016multi} & 71.19 & 56.60 & 55.30 & 86.18& 77.32& 76.33\\ 
    F-PointNet (LiDAR) \cite{qi2017frustum} & 69.50 & 62.30 & 59.73 & - & - & - \\ 
    PIXOR \cite{Yang_2018_CVPR} & - & - & - & 86.79 & 80.75 & 76.60 \\ 
    VoxelNet \cite{zhou2017voxelnet} & 81.97 & 65.46 & 62.85 & 89.60 & 84.81 & 78.57 \\ \hline
    Baseline & 71.50 &63.71 &57.31 & 86.33 & 76.44 &69.72 \\ 
	\textit{Ours} & 78.81 & 65.89 & 65.19 & 87.03 & 77.15 & 76.95 \\ \hline
    \end{tabular}}
    \caption{Comparison with state-of-the-art methods on KITTI \textit{val} set.}\label{tab:val_results}
    \vspace{-0.7em}
 \end{table}

\subsection{Comparison with State-of-the-art Methods}\label{subsec:exp_comparison}
We first compare the performance of our proposed network (\textit{Ours}) with the baseline method which does not explicitly model aleatoric uncertainties, as well as other top-performing methods (see Tab.~\ref{tab:test_results} and Tab.~\ref{tab:val_results}). Compared to the baseline method, we only add auxiliary output layers to regress aleatoric uncertainties, without further tuning any hyper-parameters. For a fair comparison, we only consider \textit{LiDAR-only} methods. We use the 3D Average Precision for 3D detection ($AP_{3D})$ and the bird's eye view Average Precision for 3D localization ($AP_{BEV}$). The AP values are calculated at the Intersection Over Union IOU=0.7 threshold introduced in~\cite{geiger2012we} unless mentioned otherwise. 

Tab.~\ref{tab:test_results} shows the performance on KITTI \textit{test} set. We compare aginst 3D FCN~\cite{li20163d}, MV3D(BV+FV)~\cite{chen2016multi} (using birds' eye view and front view LiDAR representation as inputs), PIXOR~\cite{Yang_2018_CVPR} and VoxelNet~\cite{zhou2017voxelnet}. In the table, we can see that the baseline method performs similarly to the MV3D(BV+FV) network. By leveraging aleatoric uncertainties, our method largely improves the baseline method up to $7.15\%$ in $AP_{3D}$, while achieving on-par performance in $AP_{BEV}$. Furthermore, our method produces results comparable to PIXOR~\cite{Yang_2018_CVPR} and VoxelNet~\cite{zhou2017voxelnet} which are strong state-of-the-art methods on the KITTI leaderboard. Tab.~\ref{tab:val_results} shows the detection performance on KITTI \textit{val} set with the same train-val split introduced in~\cite{chen2016multi}. Besides PIXOR and VoxelNet, we include VeloFCN~\cite{li2016vehicle} and Frustum PointNet (LiDAR)~\cite{qi2017frustum} for comparison. Unlike the original Frustum PointNet which takes 2D region proposals predicted by a RGB image detector, we use LiDAR BEV feature maps to generate proposals, as described in~\cite{qi2017frustum}. Tab.~\ref{tab:val_results} illustrates that our method improves the baseline method up to nearly $9\%$ by modeling aleatoric uncertainties. It also outperforms all other methods in $AP_{3D}$ for the moderate and hard settings. It is noteworthy to mention that both PIXOR and VoxelNet use the data augmentation tricks to enrich the training data and prevent over-fitting. For example, VoxelNet employs perturbation, global scaling and global rotation when training the network~\cite{zhou2017voxelnet}. In contrast, we do not employ any data augmentation. The performance gain is purely from modeling the aleatoric uncertainties.  We expect further improvements if we augment the training data during the training process. To sum up, the experiments show that we can leverage aleatoric uncertainties to boost the detection performance.

\subsection{Ablation Study} \label{subsec:exp_ablation_study}
We then conduct an extensive study regarding on where to model the uncertainties, the network speed and memory, as well as a qualitative analysis. Here, all networks are trained on the \textit{train} set and evaluated on the \textit{val} set.

\subsubsection{Where to Model Aleatoric Uncertainties}
In this experiment we study the effectiveness of modeling aleatoric uncertainties in different sub-networks of our LiDAR 3D object detector. In this regard, we train another two detectors that only capture uncertainties either in RPN (Aleatoric RPN) or FRH (Aleatoric FRH), and compare their 3D detection performance with the baseline method (Baseline) and our proposed method (\textit{Ours}) that models uncertainties in both RPN and FRH. Tab.~\ref{tab:ablation_test1} illustrates the AP values and their improvements for Baseline on KITTI easy, moderate, and hard settings. We find that modeling the aleatoric uncertainties in either RPN, FRH or both can improve the detection performance, while modeling the uncertainties in both networks brings the largest performance gain in moderate and hard settings. Furthermore, we evaluate the detection performance on different LiDAR ranges, shown by Tab.~\ref{tab:ablation_test2}. Again, our method consistently improves the detection performance compared to Baseline. Modeling the uncertainties in both RPN and FRH shows the highest improvement between ranges $20-70$ meters, indicating that the aleatoric uncertainties in both networks can compensate each other. As we will demonstrate in the following section (Sec.~\ref{subsec:experiment_understanding}), our proposed network handles cars from easy, moderate, hard, near-range or long-range settings differently. It learns to adapt to noisy data, resulting in improved detection performance.  

\begin{table}[tbp]
\centering
    \resizebox{0.8\linewidth}{!}{\begin{tabular}{|c|c c c|}
    \hline
  \multirow{2}{*}{Method} & \multicolumn{3}{|c|}{$AP_{3D}(\%)$} \\ \cline{2-4}
     & Easy & Moderate & Hard \\ \hline
    Baseline &71.50 &63.71 &57.31 \\
	Aleatoric RPN &72.92 (+1.42) & 63.84 (+0.13) & 58.61 (+1.30) \\
	Aleatoric FRH & 81.07 (\textbf{+9.57}) & 65.51 (+1.80) & 65.09 (+7.78) \\
	\textit{Ours} & 78.81 (+7.31) & 65.89 (\textbf{+2.18}) & 65.19 (\textbf{+7.88})  \\ \hline
    \end{tabular}}
    \caption{Ablation study - Comparison of 3D car detection performance for easy, moderate, and hard settings on KITTI \textit{val} set. We mark the highest performance gain in bold.}\label{tab:ablation_test1}
    \vspace{-0.7em}
 \end{table}

 \begin{table}[tbp]
\centering
    \resizebox{1\linewidth}{!}{\begin{tabular}{|c|c c c c|}
    \hline
  \multirow{2}{*}{Method} & \multicolumn{4}{|c|}{$AP_{3D}(\%)$} \\ \cline{2-5}
     & 0-20 (m) & 20-35 (m) & 35-50 (m) & 50-70 (m) \\ \hline 
    Baseline &72.42 & 78.96 & 57.87 & 26.17 \\
	Aleatoric RPN &80.86 (\textbf{+8.44}) & 79.72 (+0.76) & 65.44 (+7.57) & 30.54 (+4.37) \\
	Aleatoric FRH & 79.10 (+6.68) & 83.89 (+4.93) & 61.98 (+4.11) & 29.67 (+3.50) \\
	\textit{Ours} & 80.78 (+8.36) & 84.75 (\textbf{+5.79}) & 66.81 (\textbf{+8.94}) & 34.07 (\textbf{+7.90})\\ \hline
    \end{tabular}}
    \caption{Ablation study - Comparison of 3D car detection performance for data within different distances from KITTI \textit{val} set. We use IOU=0.5 threshold in this experiment. We mark the highest performance gain in bold.}\label{tab:ablation_test2}
    \vspace{-0.7em}
 \end{table}


\subsubsection{Runtime and Number of Parameters}
We use the runtime and the number of parameters to evaluate the computational efficiency and the memory requirement. Tab.~\ref{tab:runtime} shows the results of our method relative to the baseline network. We only need additional $2\ ms$ and $26112$ parameters to predict aleatoric uncertainties during the inference, showing the high efficiency of our proposed method. 
 \begin{table}[htbp]
\centering
    \resizebox{1\linewidth}{!}{\begin{tabular}{|c|c|c c|}
    \hline
  \multirow{2}{*}{Method} &  \multirow{2}{*}{Number of parameters} & \multicolumn{2}{|c|}{Runtime} \\ \cline{3-4}
     &  & Data pre-processing & Inference \\ \hline 
    Baseline & $36192171$ & $ 97\ ms$ & $70\ ms$  \\
	\textit{Ours} & +26112 (+0.07\%) & $+0\ ms$ & $+2\ ms$ (+2.86\%) \\ \hline
    \end{tabular}}
    \caption{A comparison of the runtime and the number of parameters. The runtime is measured by averaging the predictions over all KITTI \textit{val} set on a TITAN X GPU.}\label{tab:runtime}
    \vspace{-0.7em}
 \end{table}

\subsubsection{Qualitative Analysis}
\begin{figure*}[tpb]
    \centering
    \begin{minipage}{0.96\textwidth}
	\centering
	\subfigure[]{\label{fig:det_361}\includegraphics[width=0.24\textwidth]{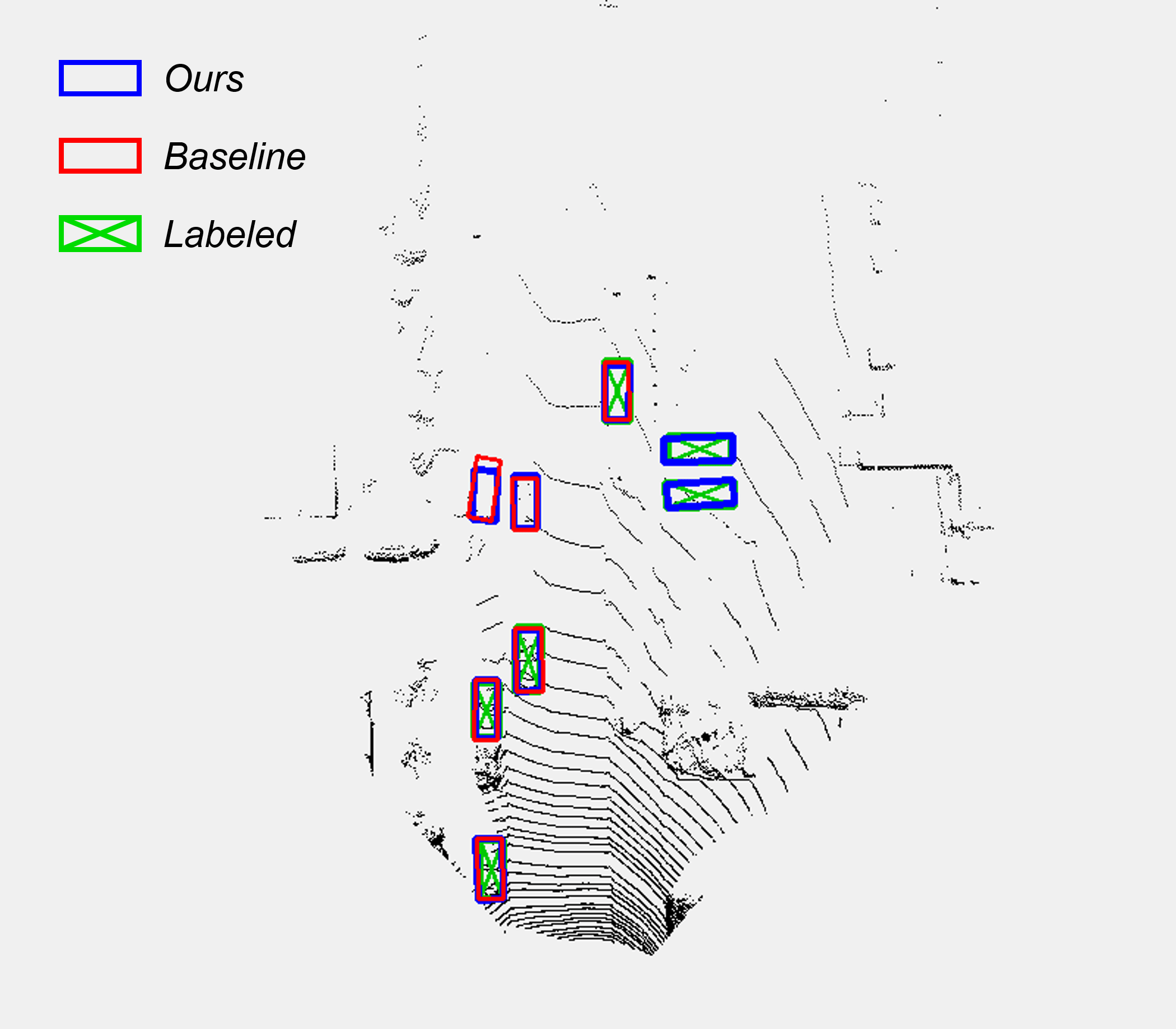}}
	\subfigure[]{\label{fig:det_658}\includegraphics[width=0.24\textwidth]{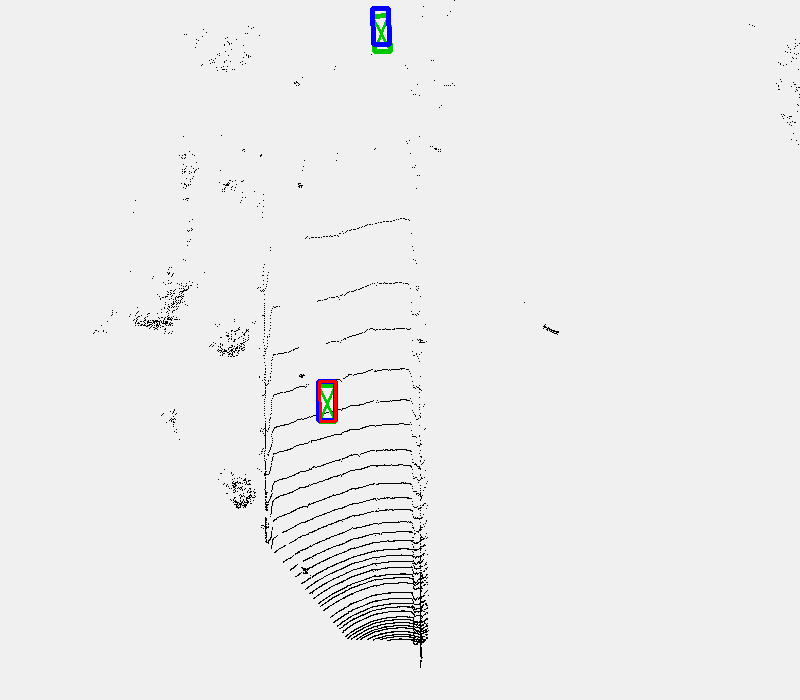}}
        \subfigure[]{\label{fig:det_169}\includegraphics[width=0.24\textwidth]{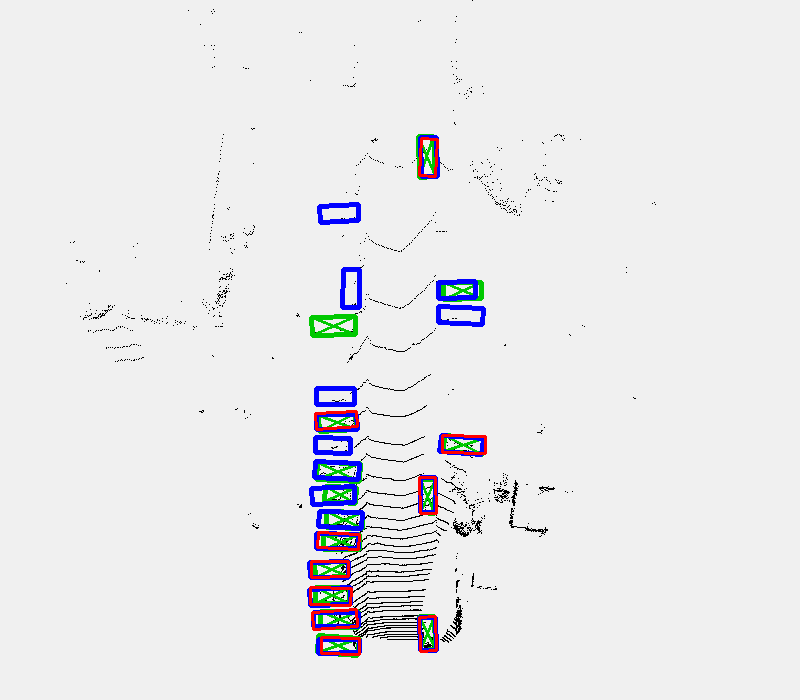}}
    \subfigure[]{\label{fig:det_323}\includegraphics[width=0.24\textwidth]{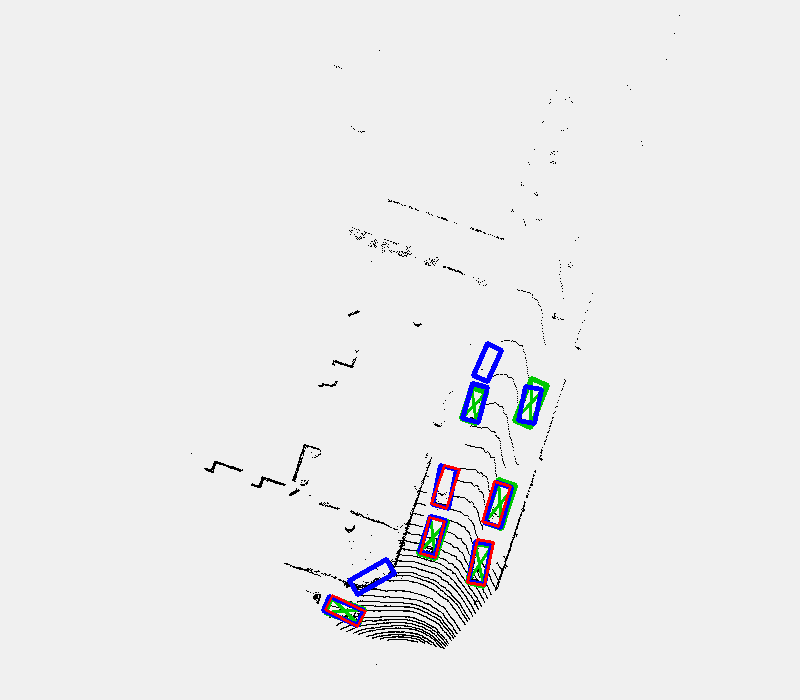}}

	\caption{Some car detection examples on KITTI \textit{val} set within the range $length \times width = [0,70] \times [$-$40,40]$ meters. The LiDAR points which are out of camera front view are not evaluated. The predictions from our proposed network (\textit{Ours}) are visualized in blue, the predictions from baseline method in red, and the labeled samples in green. Better view by magnificence.}\label{fig:detection}
\end{minipage}
\vspace{-1.0em}
\end{figure*}

Fig.~\ref{fig:detection} shows some exemplary results. In general, our proposed method can detect cars with higher recall, especially at long distance (e.g. Fig.~\ref{fig:det_658}) or in highly-occluded regions (e.g. Fig.~\ref{fig:det_169}). However, the network also tends to mis-classify objects with car-like shapes, e.g. in Fig.~\ref{fig:det_323} the network incorrectly predicts the fences on the bottom-left side as a car. Such failures could be avoided by fusing the image features from vision cameras.  
 
\subsection{Understanding Aleatoric Uncertainties} \label{subsec:experiment_understanding}
We finally conduct comprehensive experiments to understand how aleatoric uncertainties behave. We use the scalar \textit{Total Variance} (TV) to quantify aleatoric uncertainties introduced in~\cite{feng2018towards}. A large TV score indicates high uncertainty. We also use Pearson Correlation Coefficient (PCC) to measure the linear correlation between the uncertainties and the factors that influence them. We study the RPN uncertainties and the FRH uncertainties. The RPN uncertainties indicate observation noises when predicting anchor positions, while the FRH uncertainties consists of the FRH location uncertainties for the bounding box regression and the FRH orientation uncertainties for heading predictions. We evaluate all predictions with a score larger than $0.5$ from the KITTI \textit{val} set unless mentioned otherwise.

\subsubsection{Relationship Between Uncertainties}
\begin{figure}[tpb]
\vspace{0.7em}
    \centering
    \begin{minipage}{0.96\linewidth}
	\centering
	\subfigure[]{\label{fig:frh_to_rpn}\includegraphics[width=0.46\textwidth]{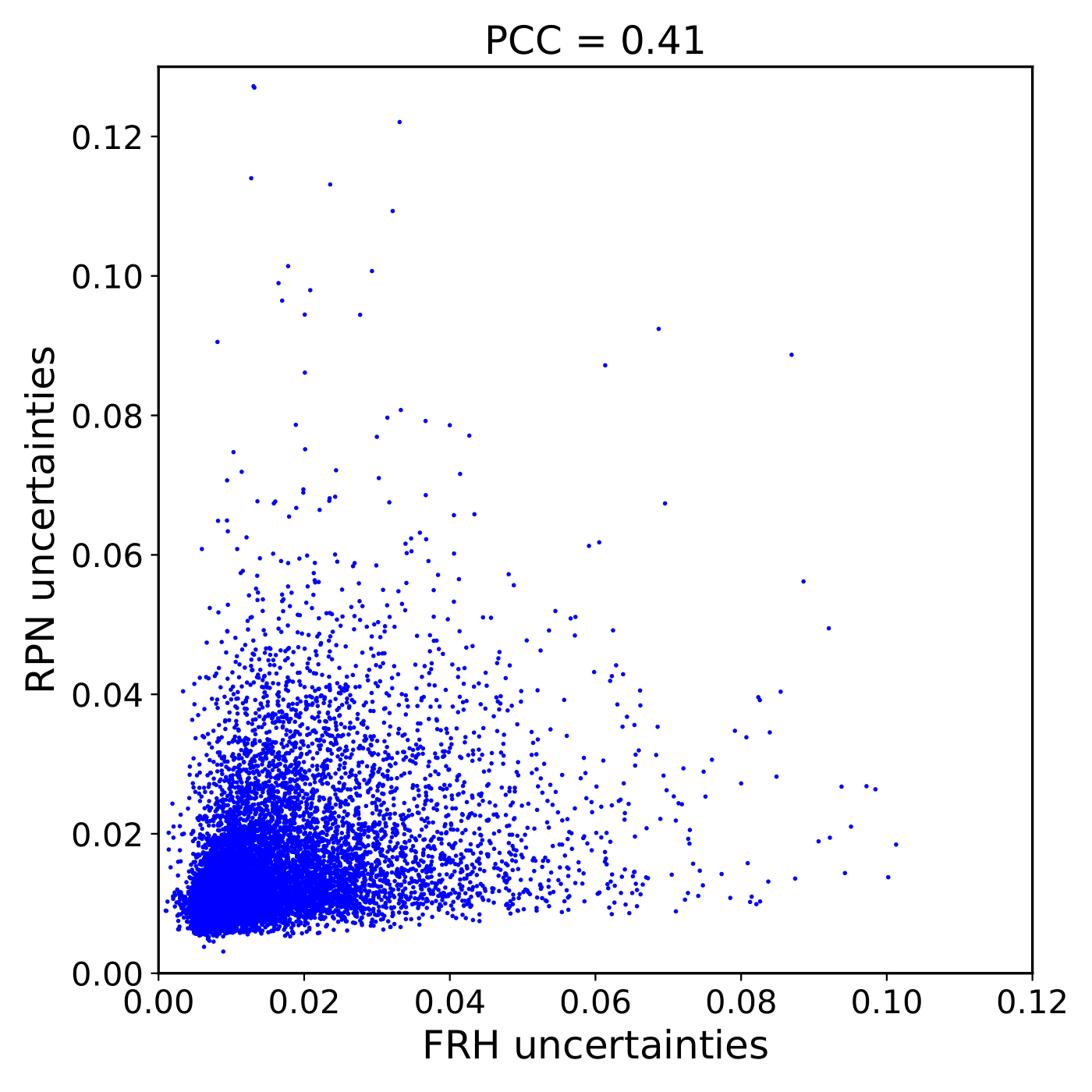}}
	\subfigure[]{\label{fig:loc_to_angle_unc}\includegraphics[width=0.46\textwidth]{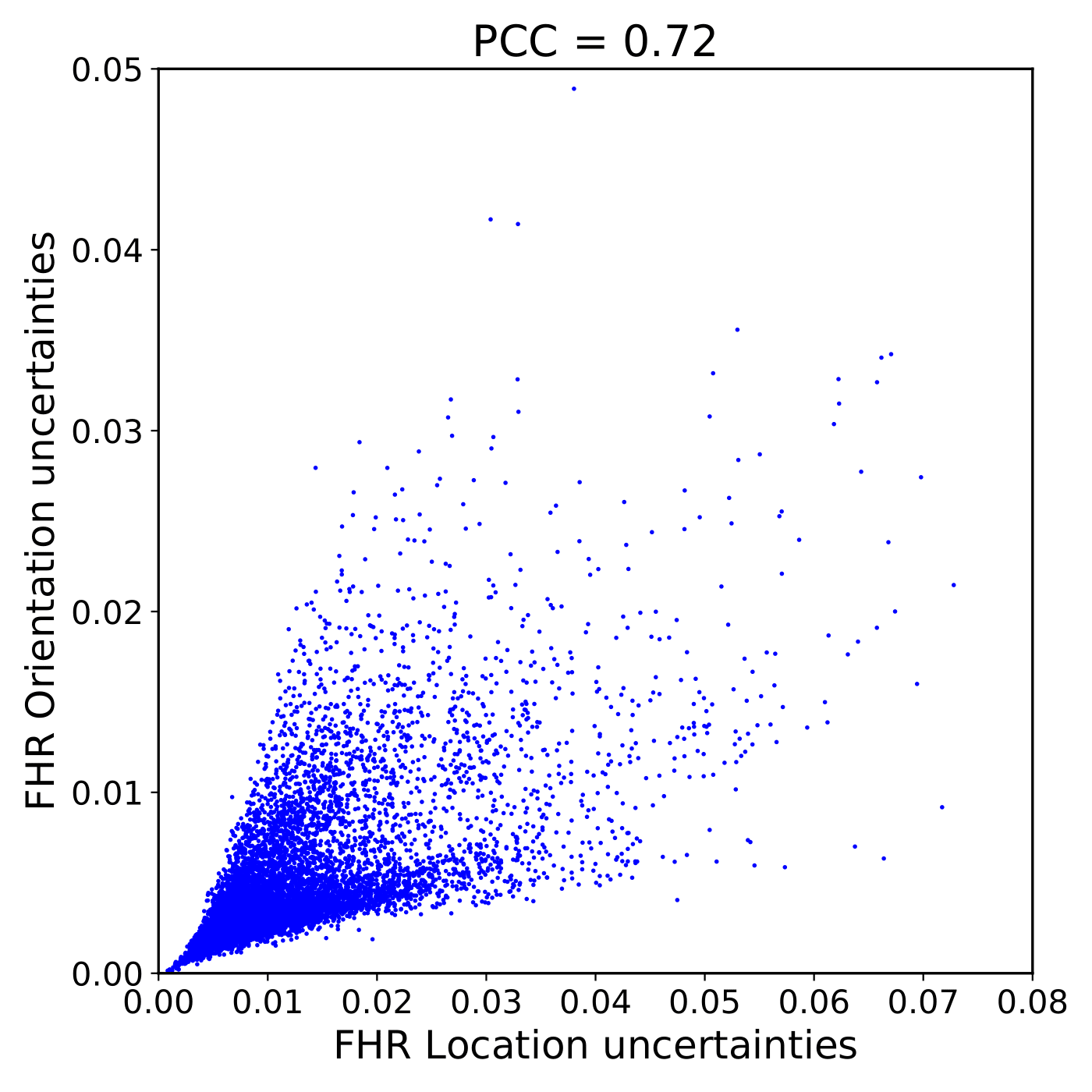}}
	\caption{(a) The distribution of uncertainties in FRH ($x$ axis) and in RPN ($y$ axis). Each scatter represents a prediction. (b) The distribution of uncertainties for FRH location ($x$ axis) and orientation prediction ($y$ axis).} 
\end{minipage}
\vspace{-1.0em}
\end{figure}

Fig.~\ref{fig:frh_to_rpn} and Fig.~\ref{fig:loc_to_angle_unc} show the prediction distribution of FRH uncertainties with RPN uncertainties, as well as FRH location uncertainties with orientation uncertainties, respectively. The uncertainties are highly correlated with each other, indicating that our detector has learned to capture similar uncertainty information at different sub-networks and prediction outputs.  

\subsubsection{Orientation Uncertainties}
\begin{figure}[tpb]
    \centering
    \begin{minipage}{0.96\linewidth}
	\centering
	\subfigure[]{\label{fig:angle_to_angle_unc}\includegraphics[width=0.5\textwidth]{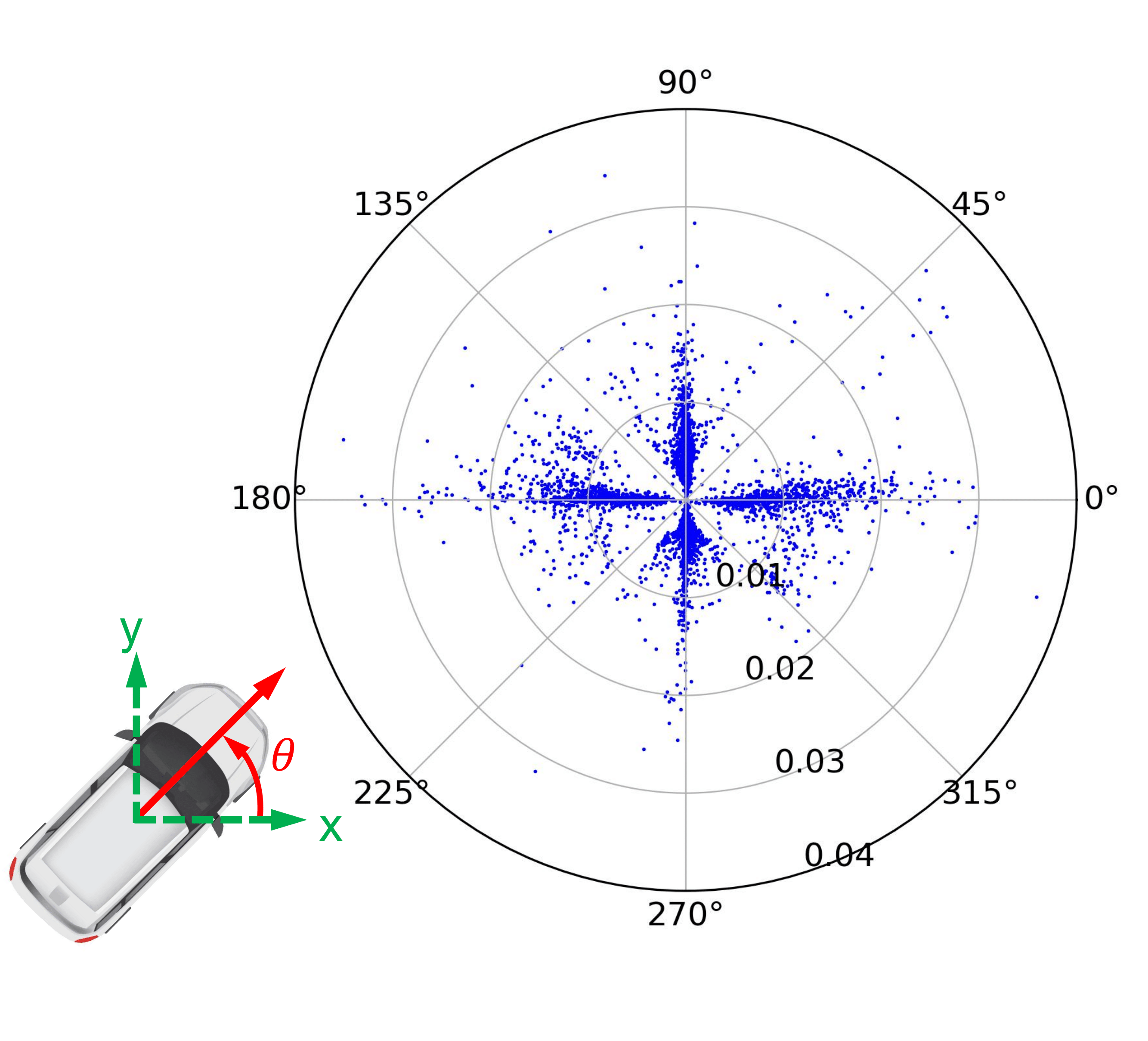}}
	\subfigure[]{\label{fig:dist_to_angle_unc}\includegraphics[width=0.46\textwidth]{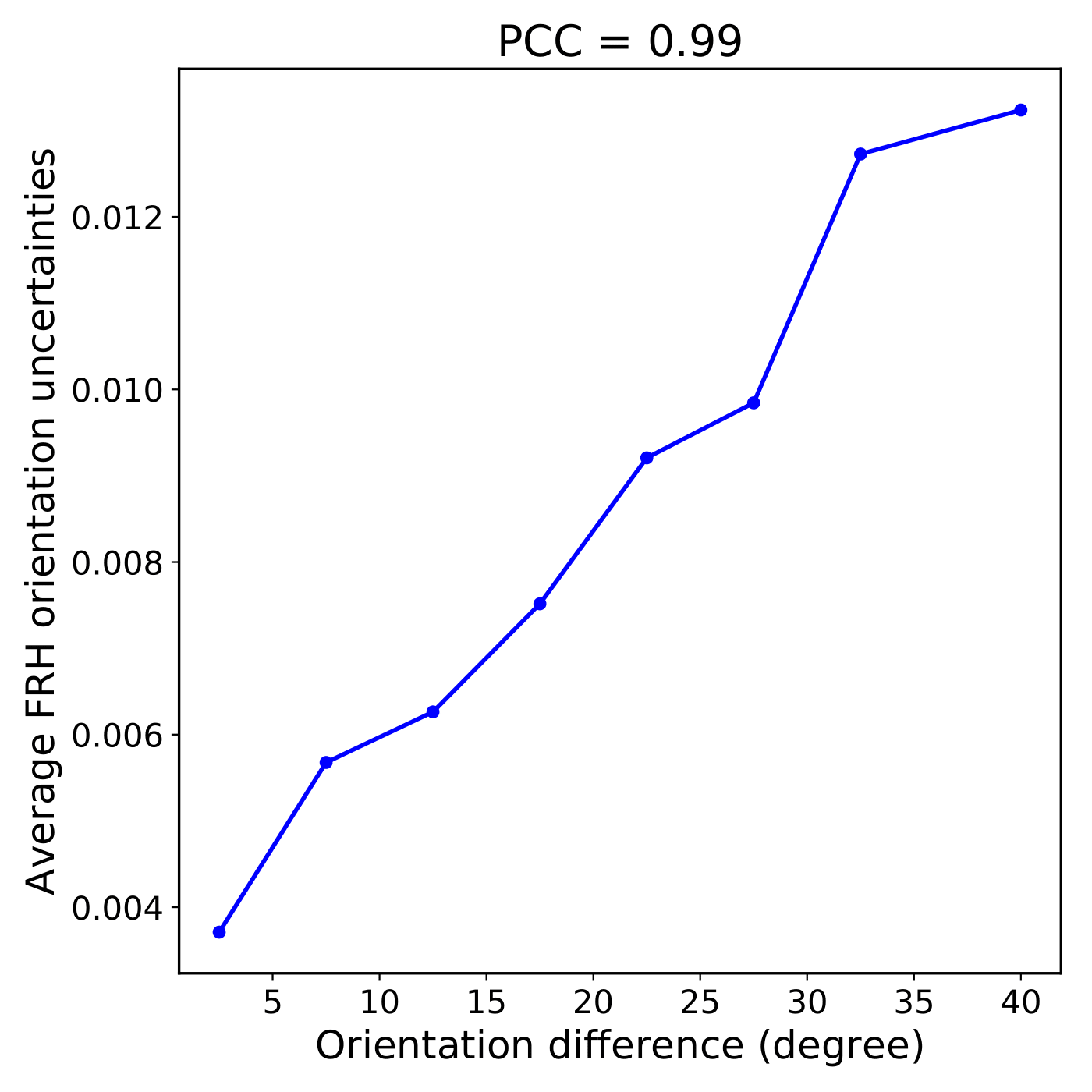}}
	\caption{(a) The prediction distribution of FRH orientation uncertainties w.r.t angle values $\theta$ (angular axis). (b) The average orientation uncertainties with orientation difference between predicted angles and the nearest base angles.}
\end{minipage}
\vspace{-1.0em}
\end{figure}

Fig.~\ref{fig:angle_to_angle_unc} illustrates the prediction distribution of FRH orientation uncertainties (radial axis) w.r.t. angle values $\theta$ (angular axis) in the polar coordinate. Most predictions lie at four base angles, i.e. $\theta=0^{\circ}, 90^{\circ}, 180^{\circ}, 270^{\circ}$. Fig.~\ref{fig:dist_to_angle_unc} shows the average orientation uncertainties with the orientation difference between the predicted angles and the nearest base angles. They are highly correlated with PCC=$0.99$, showing that the network produces high observation noises when predicting car headings that are different from the base angles. We assume that this is because most car headings in the KITTI dataset lie at four base angles, so that the network cannot be equally trained with various car headings.
\begin{figure}[tpb]
    \centering
    \begin{minipage}{0.96\linewidth}
	\centering
	\subfigure[]{\label{fig:score_to_unc}\includegraphics[width=0.46\textwidth]{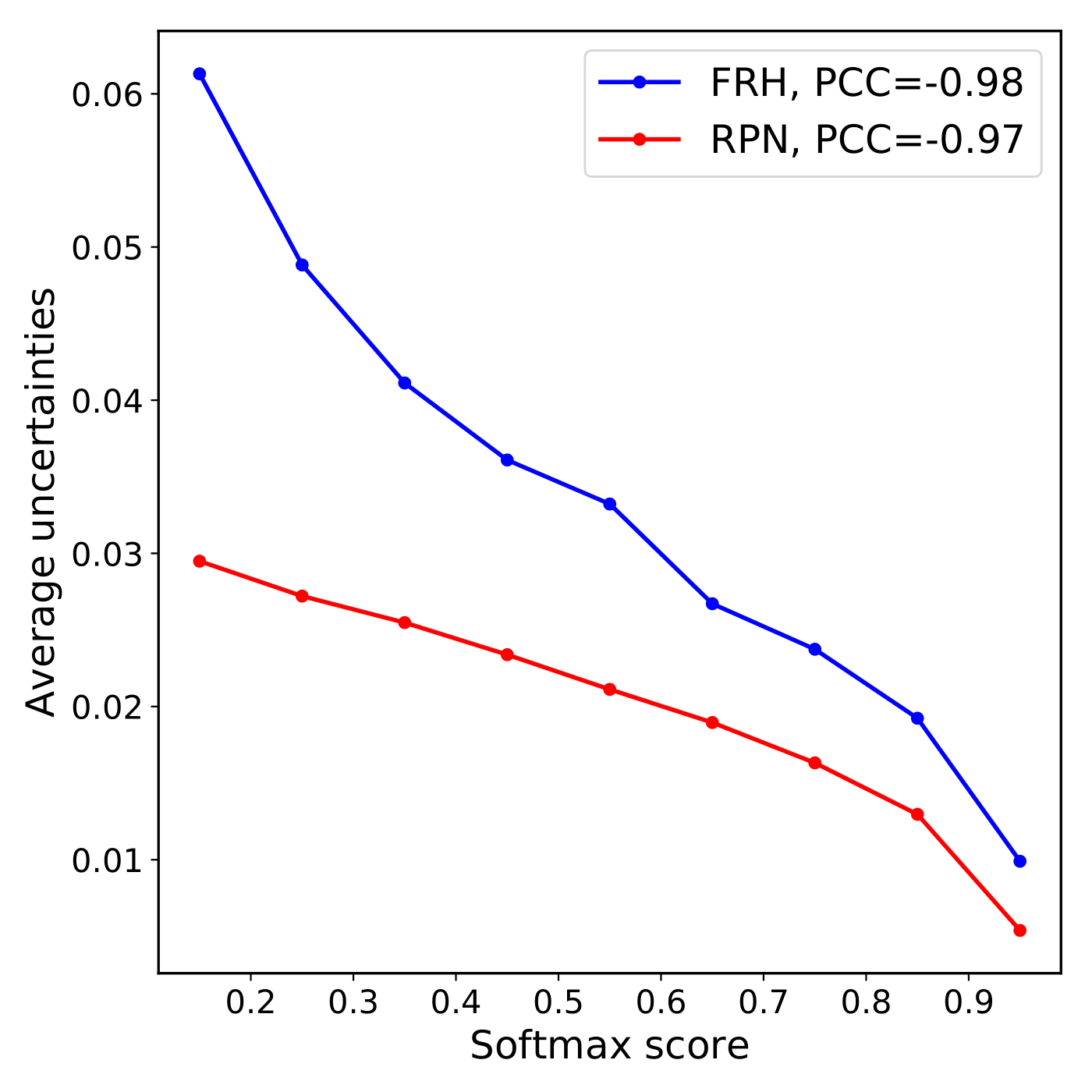}}
	\subfigure[]{\label{fig:dist_to_loc_unc}\includegraphics[width=0.46\textwidth]{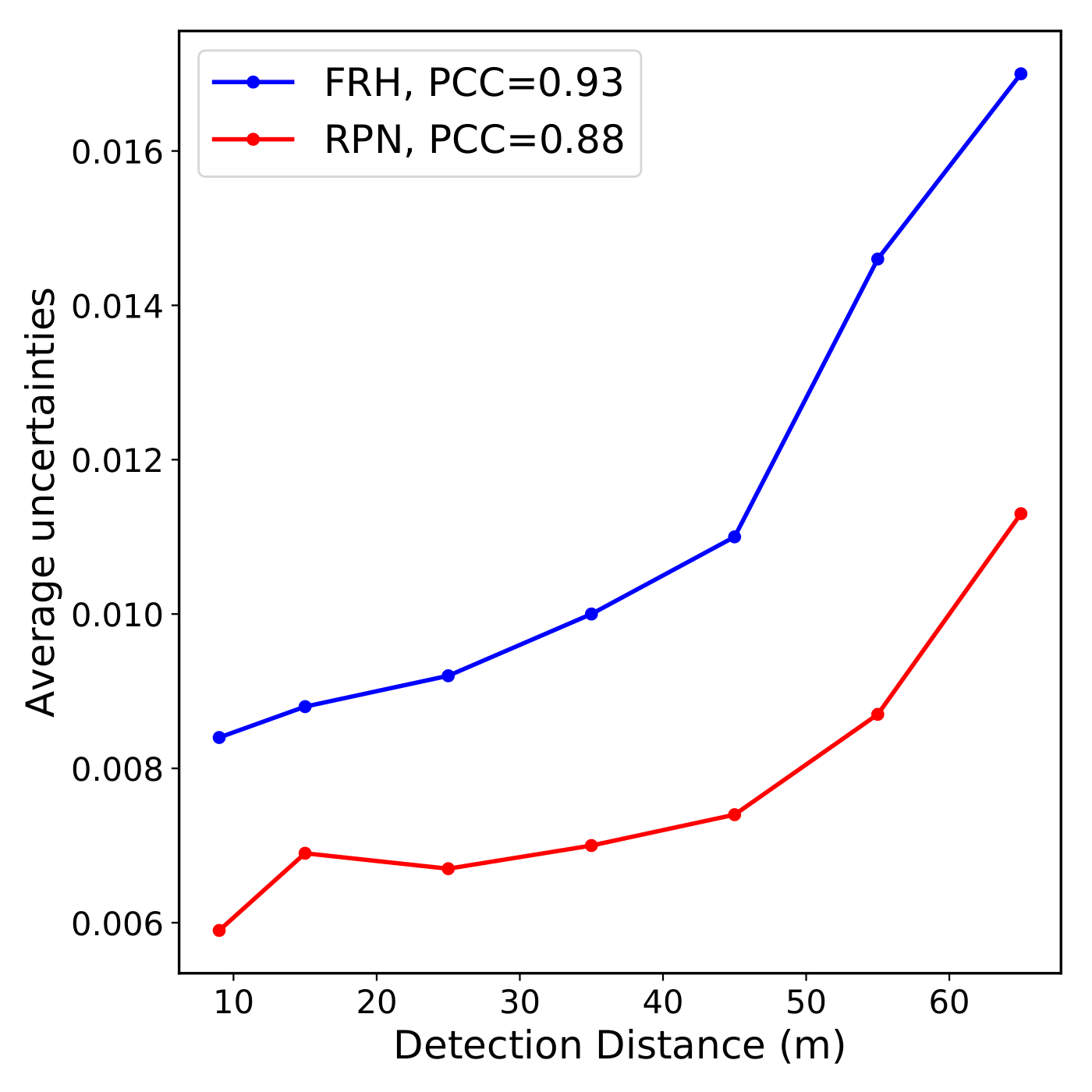}}
	\caption{(a) Average RPN and FRH uncertainties with the increasing softmax scores for anchor and final object classification. (b) Average uncertainties with the detection distance.} 
\end{minipage}
\vspace{-1.0em}
\end{figure}

\begin{figure}[tpb]
    \centering
    \begin{minipage}{0.96\linewidth}
	\centering
	\subfigure[Easy]{\label{fig:distribution0}\includegraphics[width=0.32\textwidth]{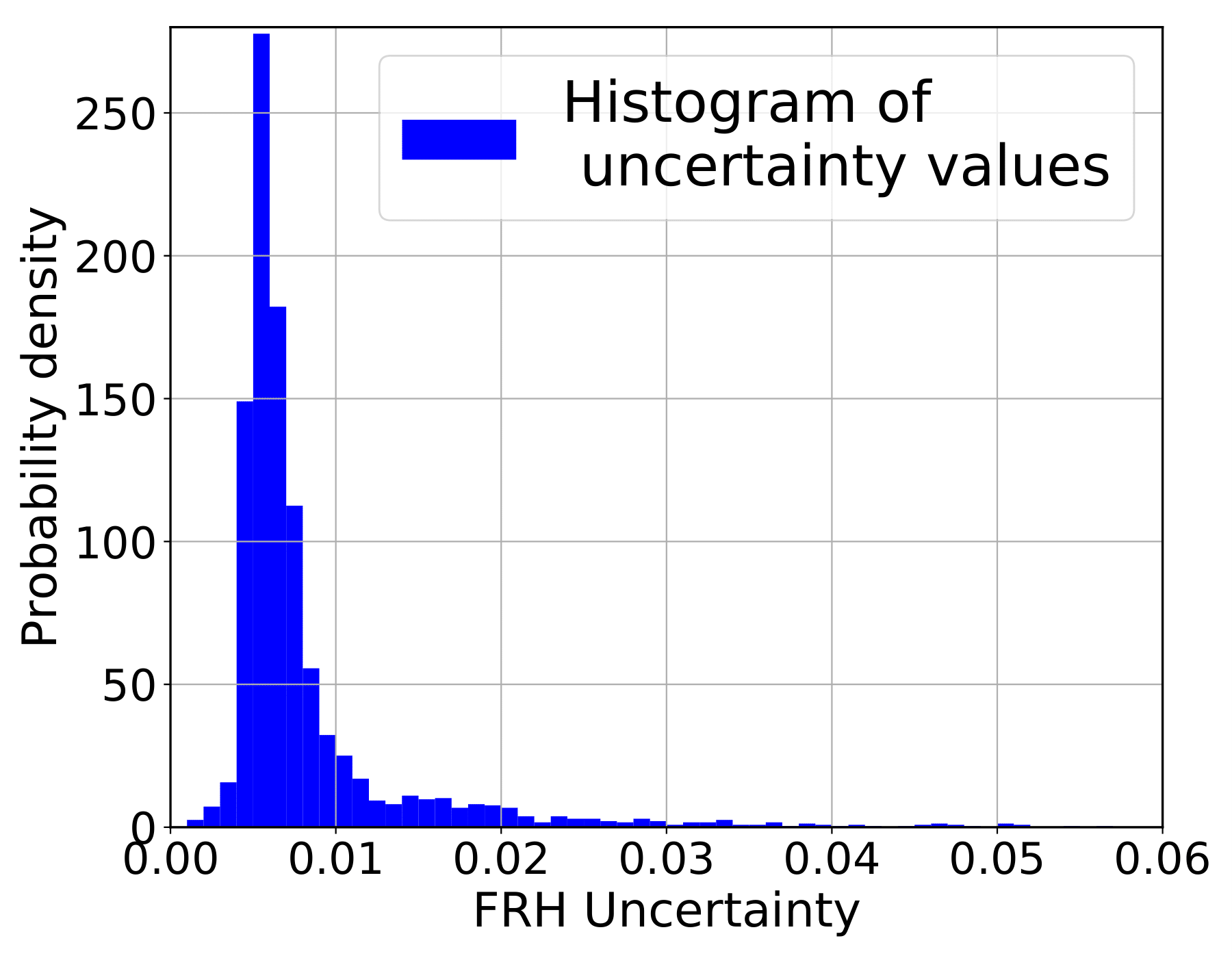}}
    \subfigure[Moderate]{\label{fig:distribution1}\includegraphics[width=0.32\textwidth]{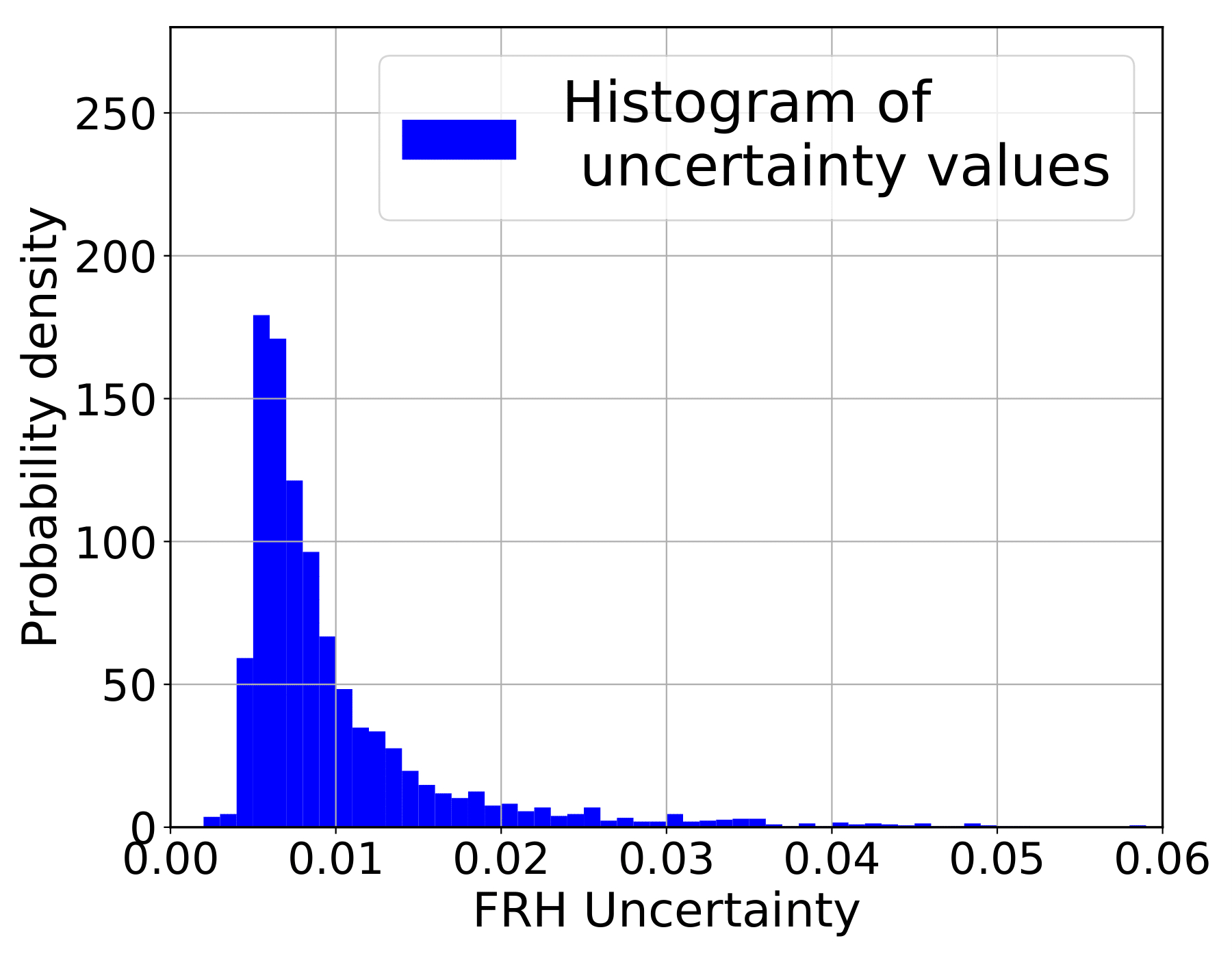}}
    \subfigure[Hard]{\label{fig:distribution2}\includegraphics[width=0.32\textwidth]{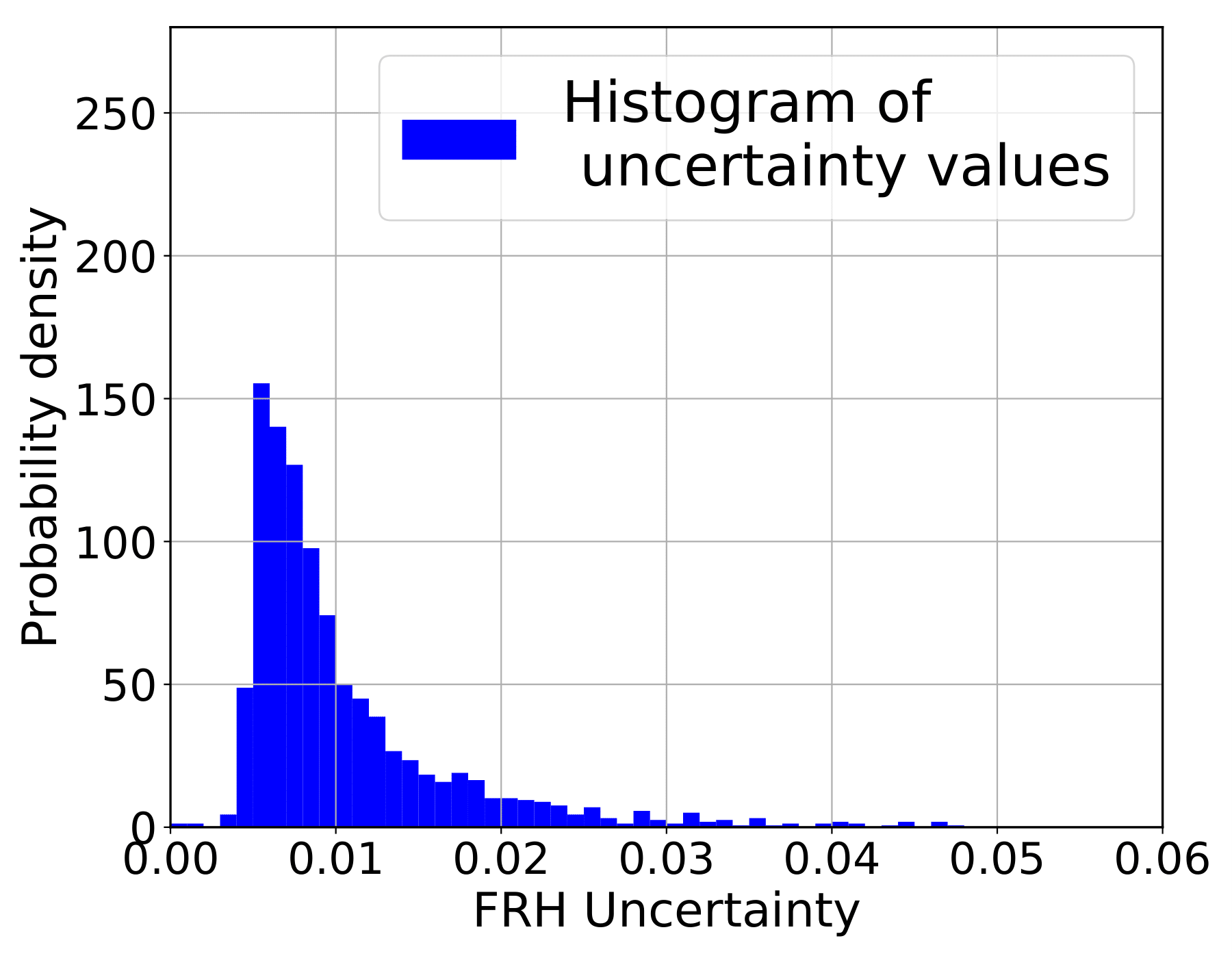}}
	\caption{Histogram of FRH uncertainties for easy, moderate, and hard settings in the KITTI \textit{val} set.} \label{fig:distribution}
\end{minipage}
\vspace{-0.6em}
\end{figure}
\begin{figure*}[tbp]
	\centering
	\begin{minipage}{1\textwidth}
		\centering
		\includegraphics[width=1\linewidth]{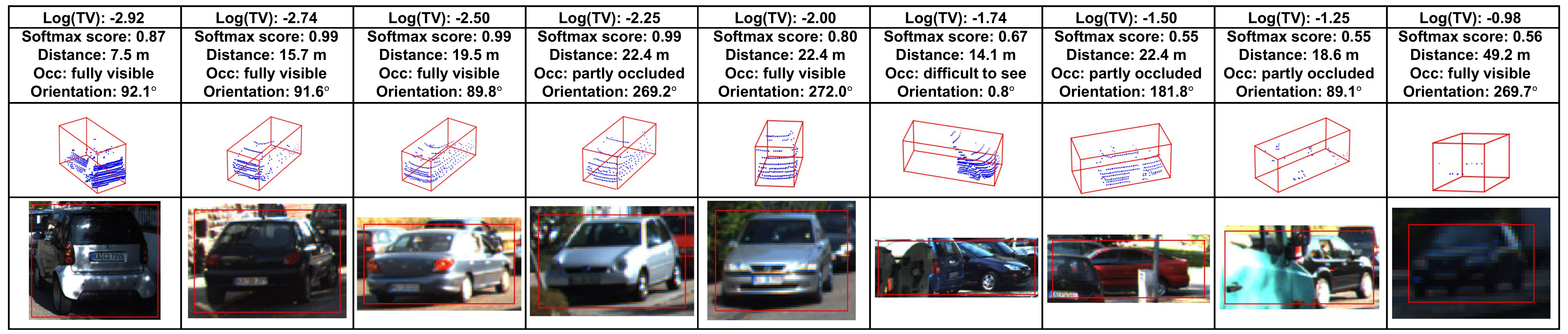}
	\end{minipage}
	\caption{Exemplary detections with the total variance of uncertainties being distributed equally at log scale ($\log(TV)$), ranging from -$3$ to -$1$. The aleatoric uncertainty is reducing from left to right. The 3D bounding boxes are predicted from our proposed network and are projected onto camera coordinate for visualization purpose only.  We also show the softmax score, object distance, level of occlusion (fully visible, partly occluded and difficult to see) as well as orientation for each detection. }\label{fig:case_study}
    \vspace{-1.0em}
\end{figure*}
\subsubsection{Relationship Between Softmax Score, Detection Distance and Uncertainties}
In Fig.~\ref{fig:score_to_unc} we plot the average RPN and FRH uncertainties with the increasing softmax scores for anchor and final object classification. We find a strong negative correlation between them. As introduced by Eq.~\ref{equ:alea_cls}, the softmax scores can be regarded as aleatoric uncertainties for classification. This indicate that our network has learned to adapt the uncertainties in regression tasks (i.e. anchor position, bounding box location and orientation) to that in the classification, i.e. the uncertainties increase as the softmax score reduces. Fig.~\ref{fig:dist_to_loc_unc} shows that the average RPN and FRH uncertainties become larger as the detection distance increases. This is because the LiDAR sensor measures fewer  reflections from a distant car, which yields high observation noises.  

\subsubsection{Uncertainty Distribution for Easy, Moderate, and Hard Settings}
We finally evaluate the FRH uncertainty distribution for easy, moderate, and hard objects, demonstrated in Fig.~\ref{fig:distribution}. The uncertainty distributions vary: for easy setting there are more predictions with lower uncertainties, whereas for hard objects which have larger occlusions, truncations, and detection distances, the network estimates higher uncertainties. The result indicates that the network has learned to treat objects from these three settings differently. 

\subsubsection{Qualitative Observations}
In Fig.~\ref{fig:case_study} we visualize nine exemplary detections whose uncertainties are equally distributed at the \textit{log scale}, ranging from -3 to -1. We observe: (1) The uncertainties are influenced by occlusion, detection distance, orientation as well as softmax score, as discussed above. (2) The network shows higher aleatoric uncertainties if there are fewer points around the car. The results show that our network has captured reasonable LiDAR observation noises.

\section{Discussion and Conclusion}
\label{sec:conclusion}
We have presented our robust LiDAR 3D object detection network that leverages heteroscedastic aleatoric uncertainties in both the Region Proposal Network and the Faster-RCNN head to significantly improve detection performance. Trained with a multi-loss function which incorporates aleatoric uncertainties, our method learns to adapt to noisy data and increases the average precision up to $9\%$, producing state-of-the-art results on KITTI object detection benchmark. Our method only requires to modify the cost function and output layers with only $2\ ms$ additional inference time. This makes our method suitable for real-time autonomous driving applications. 

We have qualitatively analyzed how the network learns to deal with noisy data in Sec.~\ref{subsec:exp_ablation_study}. The network tends to predict high uncertainties for detections that are highly occluded (Fig.~\ref{fig:case_study}), far from the ego-vehicle (Fig.~\ref{fig:dist_to_loc_unc}), with different orientations from the base angles (Fig.~\ref{fig:dist_to_angle_unc}) or with low objectness score (Fig.~\ref{fig:score_to_unc}). Therefore, our network learns less from the noisy samples during the training process by penalizing the multi-loss objective with the $\frac{1}{2\sigma^2}$ terms (Eq.~\ref{equ:objective}). Conversely, the netowrk is encouraged to learn more from the informative training samples with more LiDAR reflections. In this way, its robustness against noisy data is enhanced, resulting in improved detection performance for data from easy, moderate, hard settings (Tab.~\ref{tab:ablation_test1}), or at different distances (Tab.~\ref{tab:ablation_test2}). Note that this effect is because we explicitly model the observation noises rather than an ad-hoc solution.

Compared to the \textit{Focal Loss} \cite{Lin2017b} which incorporates prediction probability in the loss function to tackle the positive-negative sample imbalance problem, our proposed method works in an opposite way: it depreciates ``outliers" with high aleatoric uncertainties, and encourages the network to learn from the training samples with small errors. It is an interesting future work to investigate how a network behaves with the \textit{Focal Loss} and our proposed method. 

Despite that we model the aleatoric uncertainties in a two-stage object detector with LiDAR BEV input representation, our method can be extended to one-stage object detectors (such as PIXOR~\cite{Yang_2018_CVPR}), or detectors using different LiDAR representation (such as VoxelNet~\cite{zhou2017voxelnet}). In the future, we will test our method in these object detectors. We also intend to evaluate on non-motorized objects (e.g. pedestrian) using our own test vehicles and dataset.

\section*{Acknowledgment}
We thank Zhongyu Lou his suggestions and inspiring discussions. We also thank Xiao Wei for reading the script.

\bibliographystyle{IEEEtran}
\bibliography{bibliography}

\begin{thebibliography}{10}
\providecommand{\url}[1]{#1}
\csname url@samestyle\endcsname
\providecommand{\newblock}{\relax}
\providecommand{\bibinfo}[2]{#2}
\providecommand{\BIBentrySTDinterwordspacing}{\spaceskip=0pt\relax}
\providecommand{\BIBentryALTinterwordstretchfactor}{4}
\providecommand{\BIBentryALTinterwordspacing}{\spaceskip=\fontdimen2\font plus
\BIBentryALTinterwordstretchfactor\fontdimen3\font minus
  \fontdimen4\font\relax}
\providecommand{\BIBforeignlanguage}[2]{{%
\expandafter\ifx\csname l@#1\endcsname\relax
\typeout{** WARNING: IEEEtran.bst: No hyphenation pattern has been}%
\typeout{** loaded for the language `#1'. Using the pattern for}%
\typeout{** the default language instead.}%
\else
\language=\csname l@#1\endcsname
\fi
#2}}
\providecommand{\BIBdecl}{\relax}
\BIBdecl

\bibitem{ntsb_2018}
``Preliminary report highway: Hwy18mh010,'' National Transportation Safety
  Board (NTSB), Tech. Rep., 05 2018.

\bibitem{janai2017computer}
J.~Janai, F.~G{\"u}ney, A.~Behl, and A.~Geiger, ``Computer vision for
  autonomous vehicles: Problems, datasets and state-of-the-art,''
  \emph{arXiv:1704.05519 [cs.CV]}, 2017.

\bibitem{li20163d}
B.~Li, ``3d fully convolutional network for vehicle detection in point cloud,''
  in \emph{{IEEE/RSJ} Int. Conf. Intelligent Robots and Systems}, 2017, pp.
  1513--1518.

\bibitem{li2016vehicle}
B.~Li, T.~Zhang, and T.~Xia, ``Vehicle detection from 3d lidar using fully
  convolutional network,'' in \emph{Proc. Robotics: Science and Systems}, Jun.
  2016.

\bibitem{zhou2017voxelnet}
Y.~Zhou and O.~Tuzel, ``{VoxelNet}: End-to-end learning for point cloud based
  3d object detection,'' in \emph{Proc. {IEEE} Conf. Computer Vision and
  Pattern Recognition}, 2018.

\bibitem{engelcke2017vote3deep}
M.~Engelcke, D.~Rao, D.~Z. Wang, C.~H. Tong, and I.~Posner, ``{Vote3Deep}: Fast
  object detection in 3d point clouds using efficient convolutional neural
  networks,'' in \emph{{IEEE} Int. Conf. Robotics and Automation}, 2017, pp.
  1355--1361.

\bibitem{chen2016multi}
X.~Chen, H.~Ma, J.~Wan, B.~Li, and T.~Xia, ``Multi-view 3d object detection
  network for autonomous driving,'' in \emph{Proc. {IEEE} Conf. Computer Vision
  and Pattern Recognition}, 2017, pp. 6526--6534.

\bibitem{ku2017joint}
J.~Ku, M.~Mozifian, J.~Lee, A.~Harakeh, and S.~Waslander, ``Joint 3d proposal
  generation and object detection from view aggregation,'' in \emph{{IEEE/RSJ}
  Int. Conf. Intelligent Robots and Systems}, Oct. 2018, pp. 1--8.

\bibitem{xu2017pointfusion}
D.~Xu, D.~Anguelov, and A.~Jain, ``{PointFusion}: Deep sensor fusion for 3d
  bounding box estimation,'' in \emph{Proc. {IEEE} Conf. Computer Vision and
  Pattern Recognition}, 2018.

\bibitem{qi2017frustum}
C.~R. Qi, W.~Liu, C.~Wu, H.~Su, and L.~J. Guibas, ``Frustum {PointNets} for 3d
  object detection from {RGB-D} data,'' in \emph{Proc. {IEEE} Conf. Computer
  Vision and Pattern Recognition}, 2018.

\bibitem{du2018general}
X.~Du, M.~H. Ang~Jr, S.~Karaman, and D.~Rus, ``A general pipeline for 3d
  detection of vehicles,'' \emph{arXiv preprint arXiv:1803.00387}, 2018.

\bibitem{pfeuffer2018optimal}
A.~Pfeuffer and K.~Dietmayer, ``Optimal sensor data fusion architecture for
  object detection in adverse weather conditions,'' in \emph{Proceedings of
  International Conference on Information Fusion}, 2018, pp. 2592 -- 2599.

\bibitem{Gal2016Uncertainty}
Y.~Gal, ``Uncertainty in deep learning,'' Ph.D. dissertation, University of
  Cambridge, 2016.

\bibitem{qi2017pointnet}
C.~R. Qi, H.~Su, K.~Mo, and L.~J. Guibas, ``{PointNet}: Deep learning on point
  sets for 3d classification and segmentation,'' in \emph{Proc. {IEEE} Conf.
  Computer Vision and Pattern Recognition}, Jul. 2017, pp. 77--85.

\bibitem{wu2017squeezeseg}
B.~Wu, A.~Wan, X.~Yue, and K.~Keutzer, ``{SqueezeSeg}: Convolutional neural
  nets with recurrent {CRF} for real-time road-object segmentation from 3d
  lidar point cloud,'' in \emph{{IEEE} Int. Conf. Robotics and Automation}, May
  2018, pp. 1887--1893.

\bibitem{kim2016robust}
T.~Kim and J.~Ghosh, ``Robust detection of non-motorized road users using deep
  learning on optical and lidar data,'' in \emph{IEEE 19th Int. Conf.
  Intelligent Transportation Systems}, 2016, pp. 271--276.

\bibitem{schlosser2016fusing}
J.~Schlosser, C.~K. Chow, and Z.~Kira, ``Fusing lidar and images for pedestrian
  detection using convolutional neural networks,'' in \emph{{IEEE} Int. Conf.
  Robotics and Automation}, 2016, pp. 2198--2205.

\bibitem{feng2018towards}
D.~Feng, L.~Rosenbaum, and K.~Dietmayer, ``Towards safe autonomous driving:
  Capture uncertainty in the deep neural network for lidar 3d vehicle
  detection,'' in \emph{21st Int. Conf. Intelligent Transportation Systems},
  Nov. 2018, pp. 3266--3273.

\bibitem{caltagirone2017fast}
L.~Caltagirone, S.~Scheidegger, L.~Svensson, and M.~Wahde, ``Fast lidar-based
  road detection using fully convolutional neural networks,'' in \emph{{IEEE}
  Intelligent Vehicles Symp.}, 2017, pp. 1019--1024.

\bibitem{Yang_2018_CVPR}
B.~Yang, W.~Luo, and R.~Urtasun, ``{PIXOR}: Real-time 3d object detection from
  point clouds,'' in \emph{Proc. {IEEE} Conf. Computer Vision and Pattern
  Recognition}, 2018, pp. 7652--7660.

\bibitem{mackay1992practical}
D.~J.~C. MacKay, ``A practical {B}ayesian framework for backpropagation
  networks,'' \emph{Neural Computation}, vol.~4, no.~3, pp. 448--472, 1992.

\bibitem{kendall2017uncertainties}
A.~Kendall and Y.~Gal, ``What uncertainties do we need in {Bayesian} deep
  learning for computer vision?'' in \emph{Advances in Neural Information
  Processing Systems}, 2017, pp. 5574--5584.

\bibitem{hinton1993keeping}
G.~E. Hinton and D.~Van~Camp, ``Keeping the neural networks simple by
  minimizing the description length of the weights,'' in \emph{Proc. 6th Annu.
  Conf. Computational Learning Theory}.\hskip 1em plus 0.5em minus 0.4em\relax
  ACM, 1993, pp. 5--13.

\bibitem{graves2011practical}
A.~Graves, ``Practical variational inference for neural networks,'' in
  \emph{Advances in Neural Information Processing Systems}, 2011, pp.
  2348--2356.

\bibitem{gal2017deep}
Y.~Gal, R.~Islam, and Z.~Ghahramani, ``Deep bayesian active learning with image
  data,'' in \emph{International Conference on Machine Learning}, 2017, pp.
  1183--1192.

\bibitem{Beluch_2018_CVPR}
W.~H. Beluch, T.~Genewein, A.~Nürnberger, and J.~M. K\"ohler, ``The power of
  ensembles for active learning in image classification,'' in \emph{Proc.
  {IEEE} Conf. Computer Vision and Pattern Recognition}, Jun. 2018.

\bibitem{kampffmeyer2016semantic}
M.~Kampffmeyer, A.-B. Salberg, and R.~Jenssen, ``Semantic segmentation of small
  objects and modeling of uncertainty in urban remote sensing images using deep
  convolutional neural networks,'' in \emph{IEEE Conference on Computer Vision
  and Pattern Recognition Workshops}, 2016, pp. 680--688.

\bibitem{kendall2015bayesian}
A.~Kendall, V.~Badrinarayanan, and R.~Cipolla, ``Bayesian {SegNet}: Model
  uncertainty in deep convolutional encoder-decoder architectures for scene
  understanding,'' in \emph{Proc. British Machine Vision Conf.}, 2017.

\bibitem{kendall2015modelling}
A.~Kendall and R.~Cipolla, ``Modelling uncertainty in deep learning for camera
  relocalization,'' in \emph{{IEEE} Int. Conf. Robotics and Automation}, May
  2016, pp. 4762--4769.

\bibitem{miller2017dropout}
D.~Miller, L.~Nicholson, F.~Dayoub, and N.~S{\"u}nderhauf, ``Dropout sampling
  for robust object detection in open-set conditions,'' in \emph{{IEEE} Int.
  Conf. Robotics and Automation}, 2018.

\bibitem{kendall2017multi}
A.~Kendall, Y.~Gal, and R.~Cipolla, ``Multi-task learning using uncertainty to
  weigh losses for scene geometry and semantics,'' in \emph{Proc. {IEEE} Conf.
  Computer Vision and Pattern Recognition}, 2018.

\bibitem{ilg2018uncertainty}
E.~Ilg, O.~Ci{\c{c}}ek, S.~Galesso, A.~Klein, O.~Makansi, F.~Hutter, and
  T.~Brox, ``Uncertainty estimates and multi-hypotheses networks for optical
  flow,'' in \emph{European Conference on Computer Vision (ECCV)}, 2018.

\bibitem{ren2015faster}
S.~Ren, K.~He, R.~Girshick, and J.~Sun, ``{Faster R-CNN}: Towards real-time
  object detection with region proposal networks,'' in \emph{Advances in Neural
  Information Processing Systems}, 2015, pp. 91--99.

\bibitem{geiger2012we}
A.~Geiger, P.~Lenz, and R.~Urtasun, ``Are we ready for autonomous driving? the
  {KITTI} vision benchmark suite,'' in \emph{Proc. {IEEE} Conf. Computer Vision
  and Pattern Recognition}, 2012.

\bibitem{Lin2017b}
T.~Lin, P.~Goyal, R.~Girshick, K.~He, and P.~Dollár, ``Focal loss for dense
  object detection,'' in \emph{Proc. IEEE International Conference on Computer
  Vision (ICCV)}, Oct 2017, pp. 2999--3007.

\end{thebibliography}

\end{document}